\documentclass[submission,copyright,creativecommons]{eptcs}
\providecommand{\event}{CREST 2018} 
\usepackage{breakurl}             
\usepackage{underscore}           
\usepackage{mathtools}
\usepackage{microtype}

\usepackage{syntax}

\DeclareMathOperator*{\argmax}{\arg\!\max}

\title{Towards a Framework Combining\\Machine Ethics and Machine Explainability\thanks{This work is supported by the ERC Advanced Grant 695614 (POWVER) and by the Initiative for Excellence of the German federal and state governments through funding for the Saarbr\"{u}cken Graduate School of Computer Science and the DFG MMCI Cluster of Excellence.}}
\author{Kevin Baum
\institute{Universit\"{a}t des Saarlandes\\Saarland Informatics Campus\\ Saarbr\"ucken, Germany}
\institute{Department of Computer Science \\ and Department of Philosophy}
\and
Holger Hermanns\institute{Universit\"{a}t des Saarlandes\\Saarland Informatics Campus\\
Saarbr\"{u}cken, Germany}
\institute{Department of Computer Science}
\and Timo Speith 
\institute{Universit\"{a}t des Saarlandes\\Saarland Informatics Campus\\
Saarbr\"{u}cken, Germany}
\institute{Department of Computer Science \\ and Department of Philosophy}
}
\def\titlerunning{Towards a Framework Combining Machine Ethics and Machine Explainability}
\def\authorrunning{K. Baum, H. Hermanns \& T. Speith}

\usepackage{times}
\usepackage{helvet}
\usepackage{courier}
\usepackage{dsfont}

\usepackage{pgfplots, tikz}
\tikzset{>=latex}
\usetikzlibrary{shapes.multipart}

\usepackage{amsmath, amssymb}

\usepackage[utf8]{inputenc}
\usepackage{csquotes}

\usepackage{algorithm}
\usepackage[noend]{algpseudocode}

\usepackage[normalem]{ulem}

\newcommand{\powerset}{\raisebox{.15\baselineskip}{\Large\ensuremath{\wp}}}

\newcommand{\Perm}{\mathit{Perm}}

\newcommand{\dec}{\mathit{dec}}

\newcommand{\filter}{\dec_\mathit{filter}}

\newcommand{\instdec}{\mathit{dec}_\mathit{inst}}

\newcommand{\ptstr}{\mathit{force}_\mathrm{pro tanto}}
\newcommand{\oastr}{\mathit{force}_\mathrm{overall}}

\newcommand{\Arg}{\mathit{Arg}}
\newcommand{\Outcome}{\mathit{Outcome}}

\newcommand{\answer}{\mathit{AnsReq}}
\newcommand{\charge}{\mathit{Charge}}

\newcommand{\A}{\Phi}		
\newcommand{\W}{\Omega}		
\newcommand{\K}{\Theta}		
\renewcommand{\P}{\Psi}		
\newcommand{\Probs}{\Pi}	

\renewcommand{\a}{\phi}	
\newcommand{\w}{\omega}		
\newcommand{\p}{\psi}
\renewcommand{\k}{\theta}
\newcommand{\Prob}{\mathit{P}}

\newcommand{\Pfrak}{\mathfrak{P}}

\newcommand{\Supp}{\mathit{Support}}
\newcommand{\Choice}{\mathit{Choice}}

\newcommand{\EI}{E_{1,2}}
\newcommand{\EII}{E_{2,3}}

\newcommand{\rel}{\mathit{relevance}^\Pfrak}

\newcommand{\fif}{\textrm{if}}

\def\denquote#1{\lq{#1}\rq}

\newcommand{\defeq}{\coloneqq}
\newcommand{\eqdef}{\eqqcolon}

\newcommand{\tighten}{\vspace*{-3ex}}

\newtheorem{example}{Example}

\begin{document}
\maketitle

\begin{abstract}
We find ourselves surrounded by a rapidly increasing number of autonomous and semi-autonomous systems. Two grand challenges arise from this development: Machine Ethics and Machine Explainability.
Machine Ethics, on the one hand, is concerned with behavioral constraints for systems, so that morally acceptable, restricted behavior results; Machine Explainability, on the other hand, enables systems to explain their actions and argue for their decisions in a way that human users can understand and justifiably trust them.  

In this paper, we try to motivate and work towards a framework combining Machine Ethics and Machine Explainability. Starting from a toy example, we detect various desiderata of such a framework and argue why they should and how they could be incorporated in autonomous systems. Our main idea is to apply a framework of formal argumentation theory both, for decision-making under ethically motivated constraints and for the task of generating useful explanations based on these constraints given only limited knowledge of the world. The result of our deliberations can be described as a first version of an ethically motivated, principle-governed framework combining Machine Ethics and Machine Explainability. 
\end{abstract}

\section
%
{Introduction}
(Semi-)Autonomous systems are pervading the world we live in. These systems start to deeply affect our lives and we, in turn, become more and more dependent on their operations. Several important questions arise: How should machines be constrained, such that they act morally acceptably towards humans? And how should conflicts between such constraints and the traditional means-ends based decision-making be resolved? These questions concern \textit{Machine Ethics} -- the search for practically implementable behavioral constraints for systems, enabling them to exhibit morally acceptable behavior.
Although some researchers believe that implemented Machine Ethics is a sufficient precondition for humans to reasonably develop trust in autonomous systems, we want to discuss why this is not true in cases of imperfect information of the systems. We instead see the need to supplement Machine Ethics with means to enable \textit{justified trust} in autonomous systems. We argue that there is at least one feasible supplement for Machine Ethics doing the job: \textit{Machine Explainability} -- the ability of an autonomous system to explain its actions and to argue for them in a way comprehensible for humans.
In this paper we try to demonstrate how these two fields, Machine Ethics and Machine Explainability, are intertwined, and we propose the nucleus of a formal framework combining Machine Ethics and Machine Explainability. This work thereby goes beyond the rough thoughts and ideas we presented in \cite{baumMeme}.


\noindent\textbf{Related Work.}\quad
\advance\textheight14mm
Machine Ethics is becoming a serious research field, as first systematic works have been published in the last years (cf. \cite{anderson, wallach}, see also \cite{dennis} for a short overview of techniques and challenges). As James H. Moor pointed out (cf. \cite{moor}), Machine Ethics can be understood as a rather broad term, ranging from ethically motivated restrictions on the behavior of complex and possibly autonomous systems to the implementation of full-fledged moral capacities, involving deep, philosophical concepts of autonomy and deliberation, as well as free will. 
Following the systematizing ideas from \cite{wallach} regarding different degrees of moral artificial agents, we think that for now Machine Ethics should not aim directly for \emph{true ethical} decision-making. The most pressing task of Machine Ethics is rather to find a way of describing and implementing \emph{ethically constrained instrumental} decision-making.\footnote{Instrumental decision-making is means-ends oriented decision-making that tries to find the right means to achieve specific ends, where \enquote{right} here means as much as being most efficient or cost-effective. Philosophically speaking, it is a kind of \emph{instrumental rationality}.} This should allow for principle-based, unambiguous and formal guarantees that restrict the autonomous system's behavior in a way that makes the system \emph{significantly morally better}, without necessarily implementing any moral theory and still allow it to do what it was designed for. Hence, the goal is an \emph{overall morally acceptable and desirable} system that remains useful.

In contrast to Machine Ethics, Machine Explainability aims at equipping complex and autonomous systems with means to make their decisions \emph{understandable} to different groups of addressees (cf. \cite{alonso, baumTrust, hengstler, horacek, langley}) enabling a sufficient amount of transparency and perspicuity for these systems. Doing so becomes more and more urgent: For instance, the software doping cases that surfaced in the context of the VW diesel emissions scandals made obvious that the behavior of complex systems can be very hard -- if not practically impossible -- to comprehend even for experts (cf. \cite{barthe, baumDope, dargenio}). These cases make clear that explainability plays a crucial role in regard to trust and whenever it comes to accountability questions where one needs to tell apart intentional misconduct from genuine malfunction. Especially in context of autonomous systems -- which often promise positive, societal effects --, black box systems for which nobody is able to explain their decisions, predictions or behavior will plausibly lack trust in the long run. Also, many applications of computational intelligence systems -- for instance as advisors of politicians and judges -- presuppose more than naked numbers and probabilities, at least in context of liberal democracies. They need to be scrutinizable and their outputs must be justifiable at least in principle and upon request. Thus, even under the premise that the deployment of some systems is desirable from a moral point of view (thanks to their overall effects) and even if these systems would in fact behave as morally good as logically and conceptually possible (thanks to future advances in Machine Ethics): As long as people cannot justifiedly trust the systems and cannot understand the reasons for their decision, their implementations are threatened even where desirable, and they cannot be promoted with good conscience in many areas of potentially promising application. However, Machine Explainability still is a young field and especially formal frameworks supporting explanations are rare (cf. \cite{conitzer} for a simple one)
. With this paper, we take first steps towards a method to perform ethically constrained decision making -- Machine Ethics -- in a way that in itself grounds the very possibility of Machine Explainability. 
  
\section{Developing a General Framework}


In the last decades and especially in recent years, researchers have made enormous progress in the development of autonomous systems. The knowledge and the tools to create artificial agents in the sense of autonomous problem solvers and good \textit{instrumental} autonomous decisions-makers are broadly available. These agents are instrumental decision-makers insofar, as they decide \textit{instrumentally rationally} (cf. \cite{sep-rationality-instrumental}). The goal of Machine Ethics is to extend these methods such that the resulting agents not only solve their problems instrumentally well, but also in a morally acceptable way. 
That being said, it has to be admitted that, as of yet, there is not much foundational research pertaining to those topics available. For instance, we lack the possibility to spell out problems in Machine Ethics in a formal way. We especially lack a formally precise, fruitful and unambiguous way to state moral constraints and principles. As one aspect of this paper we try to undertake steps to change that. Since we are still at the beginning of this admittedly ambitious research project, we impose some restrictions for now. For instance, while we do not need to assume a deterministic evolution of the environment, we resort to a probabilistic interpretation, for instance derived from past statistical evidence. Thus, our methods are made with aleatoric uncertainty in mind. We leave the question of how to handle cases involving \textit{epistemic} uncertainty to future research.

\subsection{The World of a Medical Care Robot}

We start our discussion by describing a toy example that serves as an example context for motivating a formal framework  and bringing it to life. The example is deliberately kept simple, but sufficiently complex and general to exemplify the challenges arising with respect to Machine Ethics.

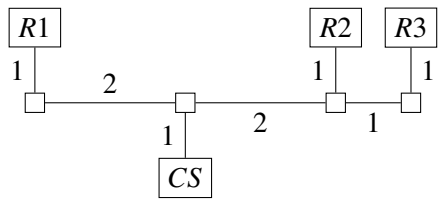
\begin{figure}[ht]
\centering
\begin{tikzpicture}
    \node[draw] at (1, 4)   (r1) {$R1$}; 
    \node[draw] at (5, 4)   (r2) {$R2$}; 
    \node[draw] at (6, 4)   (r3) {$R3$}; 
    \node[draw] at (3, 2)   (cs) {$CS$}; 
    
    \node[draw] at (1, 3)   (h1) {};
    \node[draw] at (5, 3)   (h2) {};
    \node[draw] at (6, 3)   (h3) {};
    \node[draw] at (3, 3)   (h4) {};

    \draw[] (r1) -- (h1)  node[midway, left] {$1$};
    \draw[] (r2) -- (h2)  node[midway, left] {$1$};
    \draw[] (r3) -- (h3)  node[midway, right] {$1$};
    \draw[] (cs) -- (h4)  node[midway, left] {$1$};
    
    \draw[] (h1) -- (h4)  node[midway, above, align=left] {$2$};
    \draw[] (h4) -- (h2)  node[midway, below, align=left] {$2$};
    \draw[] (h2) -- (h3)  node[midway, below, align=left] {$1$};
\end{tikzpicture}
\caption{The medical care robot's realm} \label{fig:medcare}
\end{figure}
%
\noindent We consider a medical care robot working in a hospital. There are up to three patients the robot has to take care of. Each of these patients is in a separate room ($R1$, $R2$, $R3$), and the rooms are connected by several hallways. The spatial layout of the scenario is depicted in Fig. \ref{fig:medcare}. The robot spends energy when traveling along a hallway. The energy costs depend on the distance traveled (distances are written next to the hallways). For one unit of distance, the robot needs one unit of energy. At some point the robot's battery will be depleted. To prevent this, there is  a charging station ($CS$) where the robot can recharge its battery. Once the recharging process is started, it will not stop before the battery is full again. The robot \denquote{knows} its current position and its energy level.

In our scenario, the robot listens to requests. At each point in time, each of the three patients may issue a request to the robot, asking for a task of a specific priority. Although each request has a priority when issued, this priority is deliberately \textit{not transmitted} to the robot. Instead, the robot is assumed to know appropriate and justified probability distributions regarding the tasks associated with the requests. This is necessary, as otherwise the patients could get tempted to always issue tasks of the highest priority in order to get preferential treatment. We further assume that there is only a limited number of possible tasks which can be concealed by a request. Such tasks can range from simply fetching water to doing a reanimation.
In the following we use the example to highlight a couple of central points.

\subsection{Towards a General Framework}



In this section, we work towards  a \emph{general} framework in which autonomous systems, including the above, can be described. We then extend it to be applicable to Machine Ethics in the next section. 

\tighten\paragraph{World States and (Partial) Knowledge.} We assume the autonomous system's world to be fully specifiable by assignments of a finite number of variables. Thus, a \textit{state of the world} (short: $\w$) is represented by a tuple of variables \(\w \defeq \left<\w_1, \hdots, \w_n\right>\) with corresponding domains $D_1, \hdots, D_n$. We call the set of all possible world-states $\W \subseteq \bigtimes_{i=1}^n D_i$, and we let $|\w|$ denote the number of elements in a tuple $\w$. 

At each state, the system knows some, but not necessarily all, facts about the current world state. In fact, the set of known variables may vary from state to state. For instance, our robot may know the brightness and temperature in some room if and only if it is in this room. Thus, a subset of a complete world state represents the variables the system has knowledge of. For the sake of readability, we conveniently assume the first $k$ ($1\leq k\leq n, n=|\w|$) variables $\left<\w_1, \hdots, \w_k\right>$ of a state of the world to be known variables at some given moment. We call the set of these (in general) partial world states $\K \subseteq \bigtimes_{i=1}^k D_i$. We do not rule out that the variables spanning $\K$ are dependent and, thus, the strict containment $\subsetneq$ to hold. Furthermore, we define the possible world states in the light of some knowledge $\k \in \K$ as $\W_\k \defeq \{\w \in \W ~|~ \w_{1:k} = \k\}$, where $\w_{i:j}$ for $1\leq i<j\leq n$ denotes the subtuple $\left<\w_i, \ldots, \w_j\right>$ of some tuple $\w = \left<\w_1, \ldots, \w_i, \ldots, \w_j, \ldots, \w_n\right>$.

Even if the strict inequality $k<n$ holds (as it does in most cases), we assume the system to be not clueless with regards to the remaining variables $\left<\w_{k+1}, \hdots, \w_n\right>$.  The system, hence, has justified credence regarding these variables' assignments, representable as probability distributions over the variables' domains $\Prob_{i}$: $D_{i} \rightarrow [0,1]$, $k < i\leq n$. The probability of an unknown variable having a specific value might very well be dependent on the values of known variables. So, for some $\w_i \in D_{i}$ with $k<i\leq n$ and some assignments of the known variables $\k \in \K$, we have $\Prob_{i}(\w_i) \neq \Prob_{i}(\w_i|\k)$. Call the set of all these distributions $\Probs$. 
Finally, we can assign the overall credences of the system concerning a specific configuration $\w \in \W$ as $\Prob(\w) = \prod_{i=1}^n \Prob_{i}(\w_i)$, where $\w_i$ is the $i$th variable in $\w$. As the system normally has some knowledge $\k\in\K$, we get for every $\w \in \W_\k$ the probability $\Prob(\w) = \Prob(\w|\k) = \prod_{i=k+1}^n \Prob_{i}(\w_i | \k)$.

\vspace{-0.5em}
\begin{example}
For our robot, a world state would consist of variables encoding daytime, position, energy level, energy costs for traveling, the requests in its queue and the tasks associated with it as well as some more. Since by design our robot is quite omniscient, the knowledge subset of these world states would contain everything except for the tasks associated to the requests.
\end{example}
\vspace{-0.5em}

\tighten\paragraph{Options and Actions.} 
At each decision state -- that is, in a state where the system given its context has to decide something, normally after performing an operation or triggered by some incoming event --, the system has to choose from a number of available operations. Call the operations the \textit{options} and let $\A =\{\a_1, \ldots, \a_n\}$ be the set of them. The operations available to the system will normally depend on the current state of the world, but for the sake of simplicity we do not elaborate on this dynamic here any further. Instead, we assume them to be constant over $\W$. An \textit{action} is an instantiated (i.e., chosen and performed) option and thus, by assumption, the observable decision the system has made.

\vspace{-0.5em}
\begin{example}
In case of our robot example, there are just two possible options: it can either serve the request ($\answer$) or recharge ($\charge$). 
\end{example}
\vspace{-0.5em}
\tighten\paragraph{Goal(s), Outcomes and Instrumental Decision-Making.}
We simply assume here that any system under consideration not only has at least one unambiguously defined goal, but also a method $\instdec^\Probs$ of deciding for the best means to achieve this goal (given the system's knowledge and a set of candidate options). Traditionally, the goal is to find an action that maximizes some kind of expected utility. This utility incorporates both, the uncertainty of the action's outcomes in the light of the world state's uncertainty (and possibly even some indeterminacy in the world's \denquote{reaction} to some action) as well as rewards and penalties associated with the possible outcomes.

Here is, for instance, how such an instrumental decision can be made, if we model the issue as a Markov decision problem. 
We assume that there is a function that, given the current world state, assigns to some action and another world state (i.e., a candidate for an \emph{outcome} of the action given the current world state and the action) the probability of that outcome. Formally this can be specified as $\Outcome: \W \times \A \times \W \rightarrow [0,1]$. Using the probability distributions on unknown variables (given some knowledge state) from above, we can straight-forwardly derive a function $\Outcome_\Probs: \K \times \A \times \W \rightarrow [0,1] $ 
\begin{equation*}\label{eqn:outcomesfun}
\Outcome_\Probs(\k, \a, \w) = 
	\sum_{\w' \in \W_\k} \Prob(\w'|\k)\cdot\Outcome(\w', \a, \w)  \end{equation*}
that operates on partial world states (i.e., the system's knowledge at some point). Further, let us suppose a utility function $U: \W \rightarrow \mathbb{R}$, specifying rewards and penalties as incentives for or against a specific behavior. This allows us to reformulate the need to achieve our goal as  maximization of utility.
We thus arrive at a well-posed Markov decision problem. Given some partial state of the world $\k \in \K$ representing the system's knowledge and a set of currently available options $\A$, following the standard approach to Markov decision problems, $\instdec^\Probs$ comes down to: 
\begin{equation*}\label{eqn:cullfun}
\instdec^\Probs(\A, \k) = \argmax_{\a \in \A} EU(\a | \k) \eqdef \Choice^\Probs(\A,\k) \text{, where }
\end{equation*}
$$EU(\a | \k) \defeq \sum_{\w \in \W} Outcome_\Probs(\k, \a,\w)\cdot U(\w)$$ 
is the expected utility of an option $\a \in \A$ given the system's knowledge $\k$.
Note that $\instdec^\Probs(\A, \w) \subset \A$.

Up to this point, we described a rather general class of decision problems that, as we showed, can be solved by methods associated with Markov decision problems. In the current state of our framework, we can thus, by adjusting the utilities in distinct ways, support or effectively even enforce specific decisions.

\vspace{-0.5em}
\begin{example}
Instantiating these considerations to the case of our medical care robot it seems plausible to assign rewards to the fulfilling of tasks and penalties to running out of power. Then, an execution plan which serves the most (and best rewarded) requests over the longest time possible is what we are aiming for. By setting the rewards of rescuing a person (through reanimation) higher than the penalty of exhausting the battery, we would get the result of human lives being more important than robots operating.

But is this kind of \denquote{tweakable} instrumental decision-making sufficient to get a satisfying way of making decisions in case of our robot? We believe not, because one can construct situations, in which maximizing the overall utility (for any assignment of utilities) plausibly is not what the robot ought to do. Assume, for instance, our robot is in room $R1$ and has to decide to either perform a reanimation there or to go back to the charging station. Let's assume further that the robot has enough power to reanimate, but then cannot make it back to the charging station afterwards. Assume now that with a high enough certainty, other high priority tasks -- say even other reanimations -- need to be performed later on. If our robot performs the reanimation now, it is not able to perform the other reanimations later. We can easily construct such a case in a way that makes the expected utility of charging higher than the expected utility of performing the current reanimation task.

At least some ethicists would agree that the robot \textit{ought not} to recharge now, nevertheless. It \textit{should} give preference to rescuing the life at issue at the moment of decision. But even an ethicist that does not agree with this, would likely subscribe to the claim that a robot should not be constructed in the go-recharge-way, first and foremost because of the question of \emph{trust}: Imagine that in such cases the robot would be witnessed to turn around and leave toward its charging station. People would not develop trust in that robot -- but it is important for people to trust in autonomous systems, as we already made plausible in the beginning.
So, let us suppose that, overall, the robot ought not to weight lives that way.
\end{example}
\vspace{-0.5em}

Instead, explicit and unambiguous constraints are needed that rule out some decisions and enforce others. Thus, Machine Ethics is a valid research field. This is the first central point we want to make.

\section{A Framework of Machine Ethics}


We think that we need substantially more than just instrumental decision-making as it is modeled up to now. In this section, we will give good indicators for this being the case and propose what we exactly need more.

Assume that Machine Ethics amounted to simply adjusting the utilities and disutilities in such a way that the induced behavior entirely adheres to a, say, consequentialist picture of morality,\footnote{Consequentialist theories are normative theories -- theories about the moral permissibility of actions -- that solely focus on the actions' outcomes. The consequentialist picture is driven by the idea of maximizing (moral) value and that what has  value (and disvalue) are states of affairs. Hence, what makes an action right (or wrong) is what the action changes (or promise to change) in the world. The classical source of consequentialism can be found in \cite{bentham} and a systematic discussion of it in \cite{bykvist}.} we apparently could integrate this in an instrumental decision-making procedure as sketched above. Given a full-fledged artificial system that is meant to qualify as a moral agent, and adopting such a picture of morality, adjusting the utilities, then, might very well be everything there is when it comes to implementing Machine Ethics, as such a system would take into account the effects its decisions make with regard to the question of trust as well. 

However, neither does an autonomous system qualify as a full-fledged moral agent, nor is a consequentialist picture of morality common sense.\footnote{The non-consequentialist competitors in the realm of normative ethics are the families of deontological theories (cf. \cite{kantMoral, ross}) and virtue theories (cf. \cite{anscombe, aristotle, foot, smart}). Philippa Foot prominently emphasized the tension between consequentialism and common sense (in \cite{foot}). For a recent consequentialist approach to avoid such clashes, see \cite{portmore}.} 

Therefore, the decision-making ought to be guided and \emph{restricted by explicit social and moral norms}. We, the people, want hard guarantees, forbidden actions and other desirable properties \emph{a priori}. This is what is needed for achieving the goal of Machine Ethics. This motivates the essential building block of our approach to a framework of Machine Ethics: Moral Principles.

\tighten\paragraph{Moral Principles.}
Assume it has been decided that the decision-making process of the system ought to be constrained by a number of carefully chosen, ethically motivated and ethically justified \textit{principles} $\P = \{\p_1, \hdots, \p_m\}$. Their concrete semantic interpretation will be discussed later. However, we already note that these principles are -- in line with how principles are often understood within moral philosophy (cf. \cite{dancy}) -- meant to be \textit{objective} in the sense that they are principles about which action ought (not) to be done with regards to certain states of the world. So, the question what ought to be done is determined by these principles irrespectively of the agent's information. 
We take this to be the most natural way to frame the core problem of Machine Ethics, because the behavioral restrictions one will need to implement will be of this kind, too. The restrictions themselves will be given by social or moral norms as well as by legislation, independent from the concrete design decisions and restrictions the system has.

In order to express a hierarchy of the principles, we define an order on $\P$ in two steps. First, an equivalence relation  $\approx_\P$ on $\P$ is assumed which induces $t$ equivalence classes $\P_1,\ldots, \P_t$, such that for  $1 \leq i \leq t: \p, \p' \in \P_i: \p \approx_\P \p'$. For any arbitrary principle $\p_i \in \P$, the class $[ \,\p_i\, ]$ refers to the equivalence class of $\p_i$. 
Second, we assume a strict total order $\succ_\P$ on these equivalence classes. This order is extended to the level of principles, such that for $\p, \p' \in \P: [ \,\p\, ] \succ_\P [ \,\p'\, ] \rightarrow \p \succ_\P \p'$, so as to define an overall (non-strict) weak order $\succ_\P \cup \approx_\P~\eqdef~\succeq_\P \subset \P \times \P$, thus a total preorder  on the principle set $\P$. We call $ \mathfrak{P} \defeq \left< \P, \succeq_\P\right>$ a \textit{principle structure}, giving us a hierarchy of moral principles.

Up to this point, we did not say anything about the principles' inner structure and about their content. We suggest to think of principles, in general, as functions \(\p: \W \rightarrow \powerset(\A)\) from a possible world state and the corresponding set of available options (which in context of this paper is assumed to be constantin $\W$) to a subset of these options, the set of \textit{permissible options} $\Perm^\p_{\A, \w} \subseteq \A$. Either such a principle does shrink -- or as we say \denquote{filter} -- the set of available options, then $\Perm_\p^\w \neq \A$, or it does not, and thus $\Perm^\p_{\A, \w} = \A$. We say that $\p$ has grip if and only if $\Perm^\p_{\A, \w} \neq \A$. The set of worlds $\w\in\W $ in which $\p$ has grip, namely $\{\w\in\W ~|~\Perm^\p_{\A, \w} \neq \A\} \in \powerset(\W)$, can be understood as predicate $c_\p$ and we write $\w\models c_\p$ to express that $\w \in c_\p$. Correspondingly, in this paper we represent principles as  conditionals\footnote{We want to emphasize that we do not mean to imply that principles in fact (whatever that means) need to adhere to such  a structure. We just say that for the purposes at hand principles might be \textit{modeled} as  conditionals. We leave more sophisticated models for future research.} of the form  $(c_\p \rightarrow o_\p)$ where each $o_\p$ is an \emph{option structure} $\left< \A, \succeq_\A\right>$, defined in complete analogy to the principle structures just introduced, but over the action space $\A$. Given that representation of principles, we owe the reader how to determine $\Perm^\p_{\A, \w}$ from these option structures. 

Given that some $\w \models c$ (for some $\w \in \W$ and some principle $(c \rightarrow o) \in \P$), this induces a (non-strict weak) \textit{permissibility order} $\succeq^\p_{\A, \w}$, a total preorder  on the option set $\A$. We refer to the topmost class $[ \,\a\, ]$ of this order (in the sense that $\forall \hat{\a} \in [ \,\a\, ]: \hat{\a} \succeq^\p_{\A, \w} \a'$ for all $\a' \in \A$) as $\Perm^{\p}(\A,\w) \subseteq \A$, the set of \textit{permissible options} relative to  principle $\p$ given some world state $\w$. 
The intention behind this construction is that the action to perform according to principle $\p$ in state $\w$ needs to be picked from $\Perm^{\p}(\A,\w)$, the set highest in the permissibility order associated with that principle.
Naturally, if multiple principles apply for a given state of the world
(i.e., given some world state, more than one antecedent is
fulfilled), the one highest up in the principle structure $\left< \P, \succeq_\P\right>$ is deemed decisive. But this does not exclude the probability that different principles of the same (topmost) equivalence class apply. 

Given an arbitrary set of principles $\hat{\P} \subseteq \P$ of principles and an arbitrary world state $\w \in \W$, we refer to the subset of principles which apply in this world state $\{(c \rightarrow o) \in \hat{\P} ~|~\w \models c\}$ as $\hat{\P}^\w$ and call $\hat{\P}^\w_{\max}$ the set of the maximally ranked principles in $\hat{\P}^\w$ according to $\succeq_\P$.
It seems right to identify the set of the overall permissible options $\Perm^{\hat{\P}^\w_{\max}}_{\A,\w}$ with the intersection of the permissible options according to the principles $\p \in \hat{\P}^\w_{\max}$, as this will result in the set of options permissible according to all these maximally ranked principles. Thus 
$$\Perm^{\hat{\P}^\w_{\max}}_{\A,\w} \defeq \bigcap_{\p \in \Perm^{\hat{\P}^\w_{\max}}_{\A,\w}} \Perm^\p_{\A, \w}.$$
In general, then, we can identify the set of overall permissible options in light of some principle structure $\Pfrak$, some world state $\w\in\W$ and some corresponding options space $\A$ as $\Perm^\Pfrak(\a, \w) \defeq \Perm^{\P^\w_{\max}}_{\A,\w}$. 

The question arises, whether this intersection will be guaranteed to be non-empty. The answer obviously depends on the properties we require for principles in the same equivalence class. If we require our system both, to only perform permissible options in the above defined sense and to never stop operating (and thus fulfill liveness), then we should require principles in the equivalence classes to be such that this intersection is never empty. 

This finding, however, echoes that sometimes there seemingly are principles which would result in what philosophers call \emph{true moral dilemmas}: situations where no option is permissible at all. Not all moral theories allow for such dilemmas to exist (for instance, consequentialist theories normally do not allow them). While we will not rule out that an unsatisfiable (relative to some $\w\in\W$) subset of principles of the same equivalence class may be a subset of a valid set of principles for the purposes of Machine Ethics, we will disregard the question what to do in such a situation.\footnote{Another, more permissive, way to define the set of permissible options would be to set $\Perm^\Pfrak(\A, \w)$ as the union of permissible sets regarding the principles in the highest ranked, non-empty equivalence class that have grip in $\w$\label{fn:tmd}.} We leave an axiomatization of this requirement for future research.

All this -- the principle structure and the method of finding the permissibility relation on the action -- coalesce into a function we call \emph{deontic filter}. Here, \enquote{deontic} indicates that something is about what \emph{ought} to be the case according to some standard or norm, such as a social or moral norm. 
The deontic filter thus is
$$\filter^\Pfrak(\A, \w) = \Perm^\Pfrak(\A, \w).$$ 
\vspace{-0.5em}
\begin{example}
We now apply this part of our framework to our robot example. As mentioned before, $\A \defeq \{\answer$, $\charge\}$. 
%
Below is one plausible way the robot's deontic filter could look like. Since we want to use the example just to give a vivid picture of how things would look, we spare applying the whole formalism here, but for a little more detail, see \cite{baumMeme}.
\label{sec:filter}
\begin{equation*}\label{eqn:filterfun}
\filter(\w,  \A) = {\left\lbrace
	\begin{array}{@{}l@{\;}l@{\;}c@{\;}l}
    	\{\answer\} &&,& \textrm{if the priority of the task associated to the request is high}\\ 
        &&& \textrm{and the current energy level would suffice to serve it;} \\
    	\{\charge\}&&,& \textrm{if the priority of the task associated to the request is low}\\ 
        &&&\textrm{and the current energy level would not suffice to serve it}\\ 
        &&& \textrm{and then to go back to the charging station;}\\
      \{\answer, \charge\} &&,&  \textrm{otherwise.}
    \end{array}\right.}
\end{equation*}
In order to determine which options are permissible in the light of these principles, the knowledge of the current world state is presupposed. For instance, our robot needs to know the associated task to the current request. But there were good reasons for not equipping the robot with that knowledge. 
\end{example}
\vspace{-0.5em}

Notably, deontic filtering presupposes perfect knowledge about the current state of the world. 
In full generality, this might still not be enough, since the evaluation of $\filter$ might even presuppose knowledge about de facto outcomes.
{
So, what to make out of this result? First, as we shall discuss in the sequel, given the necessary (but practically impossible) kind of full world knowledge (perfect information), the task of Machine Ethics becomes quite simple. Second, in the practically much more interesting case of incomplete knowledge (imperfect information), we must be prepared for morally suboptimal behavior (as we have already argued in \cite{baumMeme}), but also we cannot straightforwardly implement the notations introduced in this section.

Before we turn to the question of how to incorporate uncertainty in the context of deontic filtering (in Section \ref{sec:solmexpl}), we first turn to the idealized case, presenting an easy, sequential way to incorporate deontic filtering into the overall decision process in case of perfect knowledge.  
}

\subsection{An Idealized Overall Decision Pipeline}
  
We believe that the ingredients characterized above are all we need to specify a well-posed problem of Machine Ethics. But how do we then  solve such a problem operationally? 
There are two possibilities. Either, one interlocks the two decision modules $\filter^\Pfrak$ and $\instdec^\Probs$ into one. Or, one applies  $\filter^\Pfrak$ before $\instdec^\Probs$. More specifically, $\instdec^\Probs$ must be applied to the options \textit{surviving} the deontic filtering. Given that the deciding system has all it needs to evaluate $\filter^\Pfrak$, we believe the latter is more natural and easier to achieve. One just has to concatenate the $\filter^\Pfrak$ and $\instdec^\Probs$ into one larger decision procedure $\dec$, such that for each decision $\Perm^\Pfrak(\A,\w) \subset \A$ becomes the foundation of $\Choice^\Probs(\A,\k)$, rather than the full set of options $\A$. In such a sequential picture, if one wants to ensure \textit{liveness} of the system, then true moral dilemmas must be ruled out as aforementioned.

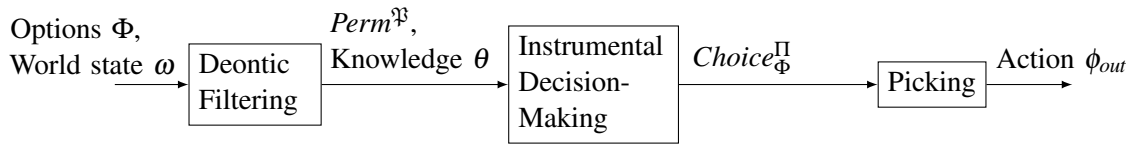
\begin{figure}[ht]
\centering
\begin{tikzpicture}
    \node[] at (1.5, 0)   (in) {}; 
    \node[draw,text width=1.5cm] at (3.5, 0)   (filter) {Deontic Filtering}; 
    \node[draw,text width=2cm] at (8, 0)   (instru) {Instrumental Decision-Making}; 
    \node[draw] at (12.5, 0)   (chooser) {Picking}; 
    \node[] at (14.5, 0)   (out) {}; 

    \draw[->] (in) -- (filter)  node[midway, above,text width=3.75cm] {Options $\A$, \\  World state $\w$};
	\draw[->] (filter) -- (instru)  node[midway, above,text width=2.25cm] { $Perm^\Pfrak$, \\ Knowledge $\k$ };
    \draw[->] (instru) -- (chooser)node[midway, above,text width=2.25cm] { $Choice^\Probs_\A$};
    \draw[->] (chooser) -- (out)  node[at end, above,text width=2cm] {Action $\a_{out}$};
\end{tikzpicture}
\caption{The sequential decision pipeline $\dec$} \label{fig:dec}
\end{figure}

Such an overall decision pipeline would consist of the following steps: \emph{deontic filtering}, \emph{instrumental decision-making} (as already introduced) and, finally and trivially, \emph{picking}: just picking a random element out of the options \denquote{surviving} the first two steps. The whole decision pipeline \(\dec\) is
visualized as flow diagram (in Fig. \ref{fig:dec}).




However, for realistic applications one needs a $\filter^{\Pfrak,\Probs}$ version of the deontic filter that operates with the de facto knowledge of the system, as $\instdec^\Probs$ does. As pointed out above, the principles encoded in the deontic filter are -- for good reasons -- objective. They only can be applied to determine the set of permissible options if one has perfect knowledge. That is, one needs full information on the complete state of the world. In other words, what we have is $\filter^\Pfrak(\A, \w)$ and what we need is $\filter^{\Pfrak,\Probs}(\A, \k)$. The framework, as it stands up to this point, thus can be understood only as a partially idealized version of what we finally long for. 
We turn to our proposal to tackle this issue in the next section.

\section{Incoporating Uncertainty \& Enabling Machine Explainability}
\label{sec:solmexpl}

\subsection{Arguments as Enablers}
How do we incorporate uncertainty in the deontic filter function? We want to present an approach that does not only solve that problem, but also enables Machine Explainability: We envision explanations as byproduct of an argument-based decision-making process.



We can easily think of the possible cases as ground for argumentation: If this or that case was true, it would give me, thanks to this or that principle, a reason for the one action rather than for the other. The probability of the cases together with the importance of the principle in question determines the strength of the resulting reason. We think that arguments can be understood as encoded reasons. And, as already proposed by Benjamin Franklin (cf. \cite{franklin}), a decision-making process (in the sense of everyday understanding of the term) can be naturally interpreted as the weighing of reasons in order to determine the right action or decision. In other words, decision-making can be thought of as an internal argumentation with pro and contra arguments for (or against) the decision or action. Furthermore, the kind of reasoning involved in our everyday decision-making seems to be non-monotonic -- further information or evidence may require the systems to retract from its decision -- and arguments are \textit{the} tool for non-monotonic reasoning as pointed out by Dung (cf. \cite{dung}).\footnote{A sophisticated framework in the context of decision-making and explanations can be found in \cite{amgoud}}

But can arguments be the right kinds of explanations? After all, there are many kinds of explanations: scientific explanations in the form of deductive-nomological models (cf. \cite{Hempel1965D}), causal explanations that relate causes with their effects, psychological explanations and many more. What we are looking for primarily are explanations that are, in terms of Davidson (\cite{davidson}), \textit{rationalizations}. These rationalizations are meant to make available to us the reasons of why the explained system decided and/or acted the way it did. Interestingly, our approach can also be seen as some kind of inductive-statistical explanation as proposed by C.G. Hempel (cf. \cite{Hempel1965I}). Given probabilities of a particular world state, we are able to derive the relative probability of each option being chosen. With this we are able to derive the robot's behavior both before and after it is witnessed. This is an important property of the kind of explanations our approach provides: a person can -- independently from the robot -- arrive at the same result.

If the system comes to its decisions based on an internal, argumentative process, the decision-making can be made transparent and rationalized in exactly the way explainability longs for. And if the argumentation-based decision-making models idealize deliberation using traditional human concepts, the obtained explanations can be expected to be \emph{comprehensible} explanations (to put it into the terms of \cite{baumTrust}: we have \emph{graspable} explanations). We, thus, suggest to resort to an \emph{argumentation-based} approach, since we believe that such an approach can not only solve the problem of finding principle-based decisions under uncertainty, but also allows to generate explanations for the resulting decisions as a byproduct.

So, provided argument-based reasoning is an appropriate approach to decision-making in the context of Machine Ethics, and arguments are the right kind of structure to encode explanations, adopting a framework of formal argumentation theory is the obvious choice for modeling and implementing these issues.\footnote{What if our robot's decision-making component is a black box, for instance, because $\instdec^\Probs$ results from some kind of machine-learning approach? Is achieving explainability hopeless then? We think that this is not necessarily so: in principle, the argumentation graphs could be derived in hindsight and maybe even \enquote{from the outside} (e.g. by some process as sketched in \cite{baumTrust}). This might come with the problem of our justifications being \emph{post hoc} rationalizations and, thus, not reflecting the \textit{true} reasons or reasoning (i.e. one needs to guarantee what \cite{baumTrust} calls \emph{accuracy}). We leave solving this problem for future research. An interesting starting point, however, could be attempts of falsification starting from inductive-statistical prognoses (cf. \cite{popper}).} Machine Explainability, then, is a \emph{byproduct} of artificial moral decision making, since the explanations are (or are extractable from) the argumentation graphs that represent what led to a decision.

\subsection{Generating an Argumentation Graph}

We propose a \emph{three step approach} for generating the argumentation graph and introduce this approach in this section. But before doing so we remark two issues: First, we shall leave out for now the question of \textit{how} to generate explanations from the argumentation graph. It will become obvious, we believe, that the (generic) arguments we describe in this section are a solid ground for spelling out explanations. Second, we suggest a procedure that is -- contrary to the above sketched approach to a framework with \textit{idealized} deontic filtering -- \textit{not} a sequential application of $\filter$ and $\instdec^\Probs$. Instead we propose a decision procedure that interlocks the two decision modules. We elaborate on that point towards the end of this section. 

We return to the three steps and the overall graph generation. They are (in their corresponding order): \emph{Case Distinction}, \emph{Reason Aggregation} and \emph{Final Action Determination}. Building upon the ingredients from the above framework -- especially $\W,\A,\Probs,EU,\Pfrak$ --, these three steps result in a \emph{bare argumentation graph} $\Gamma \defeq \left<V ,E \right >$ with $V \defeq V_1\cup V_2\cup V_3,E \defeq \EI \cup \EII$ and a couple of weight-functions
$$
\rel : V_1 \rightarrow \mathbb{R}^+,~
\ptstr :  \EI \rightarrow \mathbb{R}^+~\text{and }~
\oastr : \EII \rightarrow \mathbb{R}^+.
$$
The resulting structure $\mathfrak{G} : \left<\Gamma, ~\rel, ~\ptstr, ~\oastr\right>$ then is the \emph{complete argumentation graph}. We postpone motivating the three weight functions to the point when they are defined, but shortly introduce the elements of $\Gamma$ now. $V_i$ are the vertices (which are or contain arguments) generated in step $i$, $E$ is the set of edges representing the \emph{influences} of arguments from earlier to later steps  -- thus $\EI \subset V_1\times V_2$ and $ \EII \defeq V_2\times V_3$. 

Prior to going through the generation process step by step, we note two things: First, the whole process has to be performed for each decision that is to be made and, thus, $\mathfrak{G}$ is a function of the current knowledge $\k$ and the set of available options $\A$. Second, the process results not only in a graph, but also in a decision for a particular option. The graph is meant to be stored for the purpose of being accessible later in order to explain (or enable to explain) the corresponding decision.  

\tighten\paragraph{Step 1: Case Distinction.} As we basically aim to model something like a human agent's inner deliberative process of pondering on what she ought to do, we suggest that the systems need to take into account all possible cases $\w \in \W$ that respect the systems world-knowledge $\k \in \K$. Recall that $\W_\k \subset \W$ is the set that does so. In general, where each world state is defined by $n$ variables and $\k$ contains, by construction, the first $k$ variables of each world state  $\w$, we need to take into account $|\W_\k| \leq \prod_{i=k+1}^n |D_i|$ different world states. In order to deal with the existing uncertainty, the occurrence probabilities of the cases $\w \in \W_\k$ must be taken into account. Additionally, various cases will plausibly need to be considered not only once, but with regards to every principle $\p \in \P$ that applies to that case. We thus need an argument $\Arg_{\p}^\w$ for every $\p \in \P^\w$ of every $\w\in\W_\k$. In sum, this makes  $\prod_{\w \in \W_\k}|\P^\w|$ arguments for $V_1$. In this sense, not only the case's probability has to be considered, but also the \emph{relevance} of these applying principles, which correlates with $\succ_\P$.

First, to the form and content of the arguments. Each of these arguments  consists of three premises, logically linked by two concatenated modus ponens applications. The first premise, $P_\w$, states plainly that $\w \in \W_\k$ is the case; the second premise, $P_{\p}$, consist just of $\p$ i.e., of some $(c \rightarrow o)$; the third and final premise, $P_{\Perm_{o}}$, determines the set of permissible options according to $o$. By construction of the argument, obviously $\w \models c$, since $\p \in \P^\w$. 
Table \ref{tab:v1} shows the general form and generic content of this first layer's arguments that we now can define explicitly:
$$V_1 \defeq \{ \Arg_{\p}^\w~|~\w \in \W_\k \wedge \p \in \P^\w\}.$$
The arguments in $V_1$ only result in sets $\Perm^\p(\A,\w)$, but performing the full deontic filtering is supposed to give us $\Perm^{\Pfrak}(\A,\w)$. According to our idealized framework above, $\Perm^{\Pfrak}(\A,\w)$ is the intersection of all sets $\Perm^{\p_i}(\A,\w)$ of the principles \(\p_i\in\P_{\max}\) (i.e., the maximally ranked equivalence class with a principle that applies for $\w$). 
But this does not hold under uncertainty, because the \emph{de facto obtaining} $\w$ is unknown. For this reason we have done the case distinction in the first place. We suggest to incorporate the uncertainty in a quantitative aggregation method, respecting the \emph{probability} $\Prob(\w|\k)$ of the case (the $\w$) under consideration (given the system's knowledge $\k$) and the importance or relevance $\rel$ of the corresponding applicable principle $\p \in \P^\w$, correlating with $\succeq_\P$ given in $\Pfrak$. But how should the relevance relate to $\succeq_\P$ exactly?


Obviously, the relevance should reflect the priority ranking $\succ_\P$ over the equivalence classes $\P_1, \ldots, \P_t$ induced by $\approx_\P$. Hence, $\rel$ should be \emph{monotone} relative to $\succeq_\P$: For all $\p,\p' \in ~\left[~\p~\right]$ it holds that $\rel(\p) = \rel(\p')$ and for all $ \p,\p'$ with $\left[~\p~\right]  \succ_\P \left[~\p'~\right]$ it holds that $\rel(\p) > \rel(\p').$

However, that leaves a lot of further properties of $\rel$ unspecified. Depending on the specific properties of $\rel$ different design decisions of the reasoning process could be made. For instance, assume we want to allow that a sufficiently large number of lower ranked principles that are fulfilled (with some probability smaller than 1) can \textit{outweigh} a few higher ranked principles that are fulfilled (with the same or a lower probability). That is, we want $\rel$ to be \textit{archimedean}.

In other contexts than health care, reasoning that allows for, e.g., defeater reasons, -- roughly: reasons that silence other reasons, canceling out their strengths -- may be needed. Also, one could want to follow some kind of precautionary principle sometimes. Then a tiny chance of a principle to apply might be already enough for the system to be morally required to decide in line with this principle, no matter how improbable the case is (cf. \cite{gardiner2006} for a discussion of advantages and disadvantages of doing so). If in such cases one still wants to work with weights, the weights of specific principles might need to be infinite, so that the underlying principle is enforced. Basically, this would allow then again for \emph{true moral dilemmas} as mentioned above in context of perfect knowledge scenarios. 

So, if we would disallow any such weighing between principle fulfillments of principles in different equivalence classes, then we would want $\rel$ to map $\p$ into sets \textit{closed} under scalar multiplication (and define an order over these sets in accordance with $\succ_\P$).

Independent of how we design $\rel$ concretely, it is later assigned to each argument.

\vspace{-0.5em} 
\begin{example}
Turning one last time to our example scenario (the remaining steps are completely generic), this step is instantiated by constructing  arguments for any possible task that might be concealed by a request (since there are no other unknowns in this case). Given the original $\filter$ method from section \ref{sec:filter}, the robot knows what it is permitted to do in each possible case under consideration. Together with its probability estimates for each such case and in the light of the order of the principles, it can compute and assign all relevant aspects of this step. 
\end{example}
\vspace{-0.5em}

\tighten\paragraph{Step 2: Reason Aggregation.} In the second step, we aggregate the results from the first step. Thus, we make an argument for all the actions that \denquote{survived} the first step. Let $\Perm_{V_1}(\A) \defeq \bigcup_{\Arg_{\p}^\w \in V_1} \Perm^{\p}(\A, \w)$ be the set of all options \(\a \in \A\) for which it holds that they are permissible according to at least one applicable principle $\p\in\P$. Vice versa, let $\Supp(\a)$ be the set of arguments from $V_1$ supporting (in the sense of permitting) some option $\a\in\A$. Hence, $\Supp(\a) \defeq \{\Arg_{\p}^\w \in V_1 | \a \in \Perm^{\p}(\A, \w)\}$. For each of these options, we then need to consider what speaks in favor of it. So we start by defining: 
%
$$V_2 \defeq \{\Arg_{\a}~|~\a\in\Perm_{V_1}(\A)\}$$
We will call the output of an argument $\Arg_{\p}^\w \in V_1$ that must be taken into account into these arguments $\Arg_{\a}\in V_2$, \emph{pro tanto reasons for  option} $\a$. Consequentially
$$\EI \defeq \{\left<\Arg_{\p}^\w, \Arg_{\a}\right>~|~ \a \in \Perm^{\p}(\A, \w)\}.$$

We use $\ptstr$ as function for encoding the strength of the pro tanto reason for an option $\a$ given some argument $\Arg_\p^\w$ 
It is a function of the case's probability $\Prob(\w | \k)$ and the involved principle's  relevances $\rel(\Arg_{\p}^\w)$. 
As both quantities are weights, it seems right to aggregate them by multiplying. We leave the discussion of other kinds of aggregations, like maxing out, in special contexts for future research. We thus get:
$$\ptstr(\left<\Arg_{\p}^\w,\Arg_{\p}\right>) = \Prob(\w) \cdot \rel(\Arg_{\p}^\w)$$
Since there is no difference in strength between any two options $\a, \a' \in  \Perm^{\p}(\A, \w)$ -- that is, between two options supported by the same argument $\Arg^\w_{\p} \in V_1$ --, we will, as a shorthand, just write \enquote{$\ptstr(\Arg_{\p})$} in order to refer to that strength. 

Now we turn to the generic form and content of the arguments in $V_2$. Fundamentally different from the arguments in $V_1$, the form of the arguments in $V_2$ is dynamic: The number of premises in $\Arg_\a$ depends on the number of incoming edges, each representing a reason supporting $\a$. In other words, every $\Arg_\a \in V_2$ contains one premise for any $\Arg \in \Supp(\a)$, bringing the contributed strength with it into the argument. Additionally, one further premise is added, determining the aggregation of all these incoming reasons' strengths. So the aggregation is handled within the arguments in $V_2$. The most intuitive candidate for aggregation is simple summation of the weights. However, it may be even more controversial, whether a simple summation is the best way to aggregate reasons, than it is how to incorporate principle relevance with case probability. Answering this question (comprehensively) is clearly outside this paper's scope and, again, we leave this highly interdisciplinary question that should make use of the rich literature on that topic, to future research (cf. \cite{Horty2007, Lord2016, Mantel2017}). Table \ref{tab:v2} shows a generic version of the arguments in $V_2$.
\begin{table}[ht]
\parbox[c][6.6cm][t]{.36\linewidth}{
\centering
\begin{tabular}{l| p{3.6cm}}
\hline \multicolumn{2}{c}{Argument $\Arg_{\p}^\w$} \\
\hline\hline & \\[-1em]
(\(P_{\w}\)) & $\w$ \\[0.25em]
(\(P_{\p}\)) & \fif~$c$ then $o$ \\[0.25em]
(\(P_{\Perm_o}\)) & \fif~$o$, then $\Perm^{\p}(\A,\w)$.\\[0.25em]
\hline
\\[-0.75em]
(\(C_{i}\)) & Thus: $\Perm^{\p}(\A,\w)$. \\[0.25em]
\end{tabular}
\vspace*{-2ex}\caption{Case Distinction Arguments. The argument exemplifies the general form and generic content of the first level arguments ($V_1$). Note that by construction $\w\models c$.
}
\label{tab:v1}
}
\hfill
\parbox[c][6.6cm][t]{.685\linewidth}{
\centering
\begin{tabular}{l| p{7cm}}
\hline \multicolumn{2}{c}{Argument $\Arg_{\dec}$} \\
\hline\hline & \\[-1em]
(\(P_{\a_{i_1}}\)) & There is an overall reason supporting $\a_{i_1}$ with strength $\oastr(\a_{i_1})$. \\[0.25em]
\vdots & \vdots \\[-0.25em]
(\(P_{\a_{i_w}}\)) &  There is an overall reason supporting $\a_{i_w}$ with strength $\oastr(\a_{i_w})$. \\[0.25em]
(\(P_{\mathit{max}}\)) & Perform one randomly picked option $\a_{\emph{out}}$ of those in $\argmax_{\a \in \Perm_{V_1}(\A)} \oastr(\a)+ EU(\a |\k)$.\\[0.25em]
\hline
\\[-0.75em]
(\(C_{\mathit{final}}\)) & Thus: Perform $\a_{\emph{out}}$.\\[0.25em]
\end{tabular}
\addtocounter{table}{1}
\vspace*{-2ex}
\caption{The Final Argument. We set $w \defeq |\Perm_{V_1}(\A)|$.}
\label{tab:v3}
}

\begin{center}
\begin{tabular}{l| p{14cm}}
\hline \multicolumn{2}{c}{Argument $\Arg_{\a_i}$} \\
\hline\hline & \\[-1em]
(\(P_{i_1}\)) & There is a reason $r_1$ with strength $\ptstr(r_1) \defeq \ptstr(\Arg_{i_1})$ for $A_1$ \\[0.25em]
\vdots & \vdots \\[-0.25em]
(\(P_{i_v}\)) & There is a reason $r_v$ with strength $\ptstr(r_v) \defeq \ptstr(\Arg_{i_v})$ for $A_1$ \\[0.25em]
(\(P_{\sum}\)) & For any number of reasons $u$: If there are some reasons $r_1, \ldots, r_u$ supporting the same option $\a$ with strengths $\ptstr(r_1), \ldots, \ptstr(r_u)$, then there is an overall reason supporting $\a$ with strength $\sum_{i=1}^u \ptstr(r_i)$. \\[0.25em]
\hline
\\[-0.75em]
(\(C_{\a_i}\)) & Thus: There is an overall reason supporting $\a_i$ with strength $\sum_{\Arg\in \Supp(\a_i)} \ptstr(\Arg)$. \\[0.25em]
\end{tabular}
\addtocounter{table}{-2}
\vspace*{-2ex}\caption{Reason Aggregation Arguments: The argument exemplifies the generic form and content of the second level arguments ($V_2$). We set $v \defeq |\Supp(\a_i)|$.}
\label{tab:v2}
\end{center}
\vspace{-1em}
\end{table}

\tighten\paragraph{Step 3: Final Action Determination.} The last step is rather simple, but involves a couple of design decisions nevertheless. The remaining task, after all, is to decide for one of the options given the previous results. For this,  
we propose to combine the \textit{normative, moral force} and the \textit{normative, instrumental force} of all the remaining options. That is, we see the remaining problem as a \textit{multi-objective optimization problem} where we aim at maximizing the \emph{moral reason responsiveness} of the system on the one hand and the \textit{instrumental means-end optimality} represented by $EU$ maximization on the other. Before we elaborate on that point, first let us define the third layer of the graph.

First, $V_3$ and $\EII$. One only needs one final argument, thus 
$V_3 \defeq \{\Arg_{\dec}\}$. Since all arguments of the second level contribute to the final argument, we have $\EII \defeq V_2\times V_3$.

The final argument $\Arg_\dec$ consists of a varying number of premises, one for each $\a \in \Perm_{V_1}(\A)$, each importing the strength of the overall reason supporting $\a$. Additionally, one or more premises are included which reflect the design-decision one needs to make as mentioned above.

For the importing premises, we define the second strength function $\oastr$ (i.e., the weights for the edges from $V_2$ to $V_3$). This strength represents the aggregative normative strength of the supporting reasons in favor of each option $\a\in \Perm_{V_1}(\A)$.  These are exactly the options $\a\in\A$ for which it is possible in light of the system's current knowledge $\k\in\K$ that they are permissible according to some principle $\p= c\rightarrow o \in\P$. Here \enquote{possible} means that $\Prob(\w|\k)>0$ for some $\w \in \W_\k$ such that $\w\models c$. One could, in principle, filter out options with only tiny overall forces, for instance, in order to make the argumentation graphs computationally feasible. However, for now, we remain with the general structure as we do not see sufficient reason for such thresholding on theoretical grounds.

This time, since the aggregation of the strengths was deliberately part of the arguments in $V_2$, we only need to identify the strength of the edges with the strength in the conclusions of these arguments -- which we denote by $\Arg_{\a}.\textrm{ConStr}$\footnote{We use this (a little bit bulky) way of stating our idea in order to emphasize that the decision of how to aggregate pro tanto reasons for options is contained in the arguments in $V_2$ and is \emph{not} part of the argumentation graph generation. Whatever one plugs into the arguments has to come out as weights of the exiting edges.}:
$$\oastr\left(\left<\Arg_{\a},\Arg_{\dec}\right>\right) \defeq \Arg_{\a}.\textrm{ConStr} = \sum_{\Arg \in \Supp(\a)} \ptstr(\Arg)$$

Again, we will use a shorthand, this time \enquote{$\oastr(\a)$}, to refer to the strength supporting a specific option $\a$.
Table \ref{tab:v3} shows the generic argument $\Arg_\dec$, including the final aggregation we suggest.

Let us return to our suggestion of the simultaneous, interlocked decision method.
We start by defending our decision. We believe our approach to be superior to sequential approaches in context of the here discussed quantitative, uncertainty incorporating deontic filter method for two reasons: First, if we used a sequential approach, we could run into cases like the following. Imagine two options $\a_i$ and $\a_j$ with $\oastr(\a_i) = \oastr(\a_j) + \varepsilon$ for a negligible $\varepsilon \in \mathbb{R}^+$. Presuppose that this difference rules out $\oastr(\a_j)$ as impermissible. 
Now, imagine further that $EU(\a_i) \ll EU(\a_j)$. It seems odd that such a small difference in the supportive reasons should be decisive against an option which otherwise is much more suitable for the ends of the system. There might be filter functions operating on the reason's strengths overcoming this problem, but they will have to be more complicated and meticulously designed then a simple threshold filtering. Second, as our naming of the strength functions already indicates, we think of strengths of reasons as some kind of (normative) forces. The principles induce what could be called \emph{moral (or maybe societal) normative force}, while the instrumental design decisions encode sources of what could be called \emph{instrumental normative force}. Normative forces, in our eyes, are what pulls and pushes an agent into the direction of some option or set of options and should be combined the same way as forces are combined traditionally, namely by \textit{summation}. This justifies our decision to maximize the sum and not, for instance, the product of the two objective functions. This results in preferring some option $\a_i$ over some option $\a_j$ where $\oastr(\a_i) = 18, EU(\a_i|\k) = 3$ and $\oastr(\a_j)=10, EU(\a_j|\k) = 10$ respectively (and vice versa, for interchanged $\oastr$ and $EU$ values) in contrast to what would be the case if we decided for multiplication instead of summation. That is, we do not punish differences between the two objectives systematically. As we mentioned, we have not finally decided whether the suggestion we make here is the right one. Maybe there are good reasons for modeling this aggregation as linear combination of these forces with weights as functions of the distances of the different kinds of forces induced by some metric. For now, we stick to the simple summation and leave further deliberation, again, to future research that will need heavy involvement of Philosophy and the debate around the question of how to weigh reasons.

There is one potential practical disadvantage of our non-sequential approach. Our approach makes it rather impossible to get strong guarantees on the system's behavior. For instance, it will not be verifiable for a medical care robot (that decides using our method) that it will, whenever there is the smallest hope, attempt to rescue a life even if running out of power afterwards. There might be cases where the corresponding case is too improbable, such that the relevance of the corresponding, applicable principle is outweighed by some much more probable case in combination with a less relevant, applicable principle. 

Still, softer properties are possible and are even necessarily given by construction. Using our approach, the system would be designed such that it chooses an option that, thanks to the construction up to now, is permissible according to at least one principle applicable to at least one possible world state. Additionally, it will always choose an option that maximizes the sum of both, the combined strength of the overall reason supporting the option and the expected utility of the options given the current knowledge. In other words, it will always act upon best reasons and will be able to offer an explanation for its behavior. Maybe, our system is not \textit{verifiably} \denquote{ethical}, but it is such that its behavior always is \textit{justifiable}.

We believe this result to be the right result for many but not all contexts involving autonomous systems. For very vital or dangerous situations, so in contexts that need hard regulations -- like autonomous trading systems or even lethal autonomous weapon systems -- we, the people, demand harder guarantees. Deontic filtering then should be able to absolutely override instrumental considerations at cost of losing liveness. All this seems true to us. But now that we have defined the whole argumentation graph and finished the sketch of our framework combining Machine Ethics and Machine Explainability, we are confident that we made the corresponding \denquote{adjusting screws} evident. Our framework can thus be adapted to meet also these requirements. We believe that in this area of tension -- desired, verifiable properties on the one side and different possible design decisions on the other side -- new promising future research can be identified.

\section{Conclusion}
In this paper, we introduced a formal and general framework combining Machine Ethics and Machine Explainability. We did so in two (major) steps: first, we motivated and introduced a framework of Machine Ethics. Second, we constructed an 
instantiation of this framework enabling Machine Explainability.

In our discussion, a couple of details were left for future work. While we characterized the form and content of the arguments, we omitted a \textit{formal} characterization of their contents. Obviously, for most of the practically interesting cases, they consist of first order, modal, temporal or deontic logic formulas. This being the case, the graph needs to be supplemented by expressive means to draw conclusions: it needs a logical system, a calculus with inferences rules. Another aspect still to be explored is the space and time complexity of our approach. Additionally, we postponed a couple of optimization questions. For instance, there might be significantly fewer world states needing consideration if some variables that constitute $\w$ are dependent. Also we ignored that some variables might have very large or even uncountably infinite domains, such that considering all possible cases would be practically infeasible or even impossible. Here heuristics are needed to restrict the number of options to the most probable or important ones.

Additionally, there are at least four interesting and pressing interdisciplinary research questions left open. First there is the question of how to model the content and ordering of principles in a more sophisticated way and how to quantify these orderings -- and if one even needs to do so. After all, in light of our results, one could be inclined to switch to a framework genuinely relying on reasons if one finds a way to decouple reasons from principles (\cite{Dietrich2017} might offer a useful approach here). Second, the principle order might be context dependent. Basically, this would mean to become a particularist instead of an generalist, believing that there are \textit{no} general principles at all governing what ought to be done and that, rather, what is a normative reasons varies from context to context in an unsystematic way (cf. \cite{dancy} and \cite{sep-moral-particularism}). Third, there is a need to make decisions regarding the question of how to aggregate and weigh reasons, where the answer, as indicated before, might well be dependent on the context of application. Fourth, we have not discussed at all how to extract useful explanations of the right kind from the generated argumentation graphs.
Finally, we postponed also the question of how to handle cases involving \textit{epistemic} uncertainty (i.e., pure non-determinism) to future research as well.
There is more than enough left to work on in Machine Ethics and Machine Explainability.

\bibliographystyle{eptcs}
\bibliography{bibliography}

\end{document}